\documentclass[10pt]{article} 
\usepackage[preprint]{tmlr}
\usepackage{graphicx}
\usepackage{subcaption}
\usepackage{soul}

\usepackage{amsmath,amsfonts,bm}

\def\eqref#1{equation~\ref{#1}}

\def\1{\bm{1}}

\DeclareMathAlphabet{\mathsfit}{\encodingdefault}{\sfdefault}{m}{sl}
\SetMathAlphabet{\mathsfit}{bold}{\encodingdefault}{\sfdefault}{bx}{n}

\usepackage{hyperref}
\usepackage{url}
\usepackage{soul}

\title{Embedding Space Selection for Detecting Memorization and Fingerprinting in Generative Models}

\author{\name Jack He \email jackhe313@ucla.edu \\
      Department of Computer Science \\ University of California, Los Angeles
      \AND
      \name Jianxing Zhao \email jz2003@ucla.edu \\
      Department of Computer Science \\ University of California, Los Angeles
      \AND
      \name Andrew Bai \email andrewbai@cs.ucla.edu \\
      Department of Computer Science \\ University of California, Los Angeles
      \AND 
      \name Cho-Jui Hsieh \email
      chohsieh@cs.ucla.edu\\
      Department of Computer Science \\ University of California, Los Angeles}

\def\month{MM}  
\def\year{YYYY} 
\def\openreview{\url{https://openreview.net/forum?id=XXXX}} 

\begin{document}

\maketitle

\begin{abstract}
In the rapidly evolving landscape of artificial intelligence, generative models such as Generative Adversarial Networks (GANs) and Diffusion Models have become cornerstone technologies, driving innovation in diverse fields from art creation to healthcare. 
Despite their potential, these models face the significant challenge of data memorization, which poses risks to privacy and the integrity of generated content. 
Among various metrics of memorization detection, our study delves into the memorization scores calculated from encoder layer embeddings, which involves measuring distances between samples in the embedding spaces. 
Particularly, we find that the memorization scores calculated from layer embeddings of Vision Transformers (ViTs) show an notable trend - the latter (deeper) the layer, the less the memorization measured. It has been found that the memorization scores from the early layers' embeddings are more sensitive to low-level memorization (e.g. colors and simple patterns for an image), while those from the latter layers are more sensitive to high-level memorization (e.g. semantic meaning of an image). We also observe that, for a specific model architecture, its degree of memorization on different levels of information is unique. It can be viewed as an inherent property of the architecture.
Building upon this insight, we introduce a unique fingerprinting methodology. This method capitalizes on the unique distributions of the memorization score across different layers of ViTs, providing a novel approach to identifying models involved in generating deepfakes and malicious content. 
Our approach demonstrates a marked 30\% enhancement in identification accuracy over existing baseline methods, offering a more effective tool for combating digital misinformation.

\end{abstract}

\section{Introduction}

In the rapidly evolving field of artificial intelligence, generative models such as Generative Adversarial Networks (GANs) \citep{goodfellow2014generative} and Diffusion Models \citep{ho2020denoising} have emerged as pivotal tools, driving innovation across a myriad of domains including art \citep{ramesh2021zeroshot}, media synthesis \citep{dhariwal2021diffusion}, healthcare \citep{kazerouni2023diffusion}, and autonomous systems \citep{liu2024ddmlag}.
These models possess the remarkable capability to generate new, synthetic data instances that are often indistinguishable from real data, thereby unlocking new frontiers in data augmentation \citep{trabucco2023effective}, realistic content creation \citep{rombach2022highresolution}, etc.
However, alongside their remarkable capabilities, these models harbor the intrinsic challenge of data memorization.

Data memorization occurs when models inadvertently replicate exact or near-exact pieces of their training data \citep{somepalli2023understanding}. It presents significant concerns relating to privacy breaches, the integrity of generated content, and the perpetuation of biases ingrained in the training datasets.
The nuanced issue of memorization within generative models necessitates robust mechanisms for detecting and quantifying memorization. This mechanism can not only aid in understanding how models learn and replicate data patterns \citep{burg2021memorization} but also in implementing safeguards against misuse, such as unauthorized replication of sensitive information \citep{wang2023security}. 
Currently, the practice of measuring memorization primarily involves analyzing the embeddings generated from various layers of neural networks, such as Convolutional Neural Networks (CNNs) \citep{heusel2018gans} and Transformers \citep{pizzi2022selfsupervised}. 
However, the decision regarding which layer's embeddings should be employed for memorization measurement has not been formally standardized or thoroughly explored. 
Common practice often gravitates towards utilizing the embeddings from the penultimate layer of these networks. The lack of justification to the layer choice underscores a pressing gap in the existing methodologies, highlighting the need for a systematic exploration to identify the most effective layers for capturing memorization phenomena accurately. This paper aims to bridge this gap by conducting a comprehensive investigation into the efficacy of memorization measurement using embeddings from different layers of CNNs and transformers. We hope to establish informed guidelines for embedding layer choice, thereby enhancing the ability to correctly quantify the extent of data memorization using the existing methods of measurement.

In this research we study $C_T$-score, an embedding-based data-copying metric proposed in \citet{meehan2020nonparametric}, to measure model memorization with ViT layer embeddings. Beyond merely selecting the best layers for memorization detection, we introduce and validate a novel approach for model fingerprinting that leverages the unique distribution of the $C_T$-scores layer-wise. The idea comes from the experiment observation, that each model shows a distinctly unique trend in $C_T$-scores calculated by embeddings from different layers.
This approach is grounded in the understanding that a model's memorization characteristics are uniquely shaped by its architecture, the data it is trained on, and the optimization techniques used during training. It demonstrates superior performance relative to existing baseline methods in the task of model identification by generated images. This again highlights the vast unexplored possibility of layer-wise study of memorization scores.

\section{Related Works}
Research in data memorization has shown that deep neural networks can memorize random labels due to having significantly more learnable parameters than training examples, indicating a potential to prioritize memorization over generalization \citep{novak2018sensitivity}. 
This observation is complemented by findings from \citet{zhang2017understanding}, who demonstrated that standard regularization strategies are often inadequate in preventing such memorization. 
Furthermore, \citet{stephenson2021geometry} applied a geometric analysis method to quantify the extent of memorization, and discovered the preferentially learning for different layers.

In the realm of generative models, particular attention has been directed towards Generative Adversarial Networks (GANs) and Diffusion Models. \citet{bai2022reducing} explored methods to reduce memorization in GANs without compromising output quality. 
Concurrently, \citet{somepalli2023understanding} investigated strategies to mitigate data copying effects in Diffusion Models. 
Assessment techniques for generative models have also evolved, incorporating metrics like the popular Inception Score (IS), Frechét Inception Distance (FID) \citep{heusel2018gans}, Precision and Recall test \citep{sajjadi2018assessing}, and Self Supervised Copy Detection (SSCD) \citep{pizzi2022selfsupervised}.

Model fingerprinting aims to create a unique identifier for a machine learning model by analyzing its performance, structure, and behavior to distinguish it from other models. Studies involving model fingerprinting to trace the origins of digital artifacts have grown \citep{song2024manifpt, yu2022artificial}. Such efforts are crucial for understanding the source of generative outputs and are instrumental in the battle against the misuse of AI technologies, such as the production of deepfakes. These studies lay the groundwork for exploring more refined mechanisms to detect and mitigate memorization and to ensure the ethical use of generative models in producing digital content.

\section{Observation}
Our research uses $C_T$-score as the metric of model memorization. To calculate $C_T$-scores, generated samples are projected into an embedding space, which is often obtained from neural network encoders. Both CNN-based and transformer-based encoders are used in our initial experiment. The variance of layer-wise $C_T$-score trends resulting from the selection of encoder types is noted. More importantly, an interesting trend is consistently observed when the encoder is transformer-based.

\begin{figure}[!ht]
\begin{center}
\includegraphics[width=\linewidth]{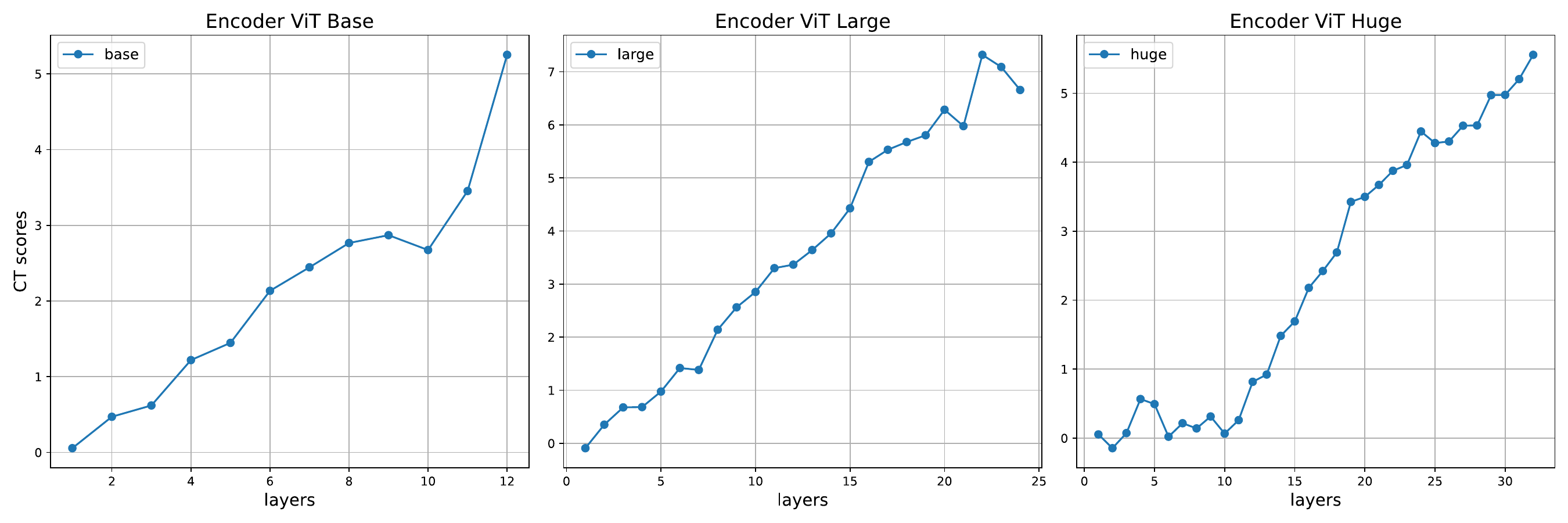}
\end{center}
\caption{\textbf{ViT encoder trend based on DDPM-generated Image}. We use 500 randomly sampled DDPM-generated \protect\citep{ho2020denoising} CIFAR-10 Images to compute the $C_T$ scores with three different Vision Transformers, namely ``vit-base-patch16'', ``vit-large-patch16'', and ``vit-huge-patch14''. We observe a consistently increasing trend.}
\label{fig:ddpm500ct}

\begin{center}
\includegraphics[width=\linewidth]{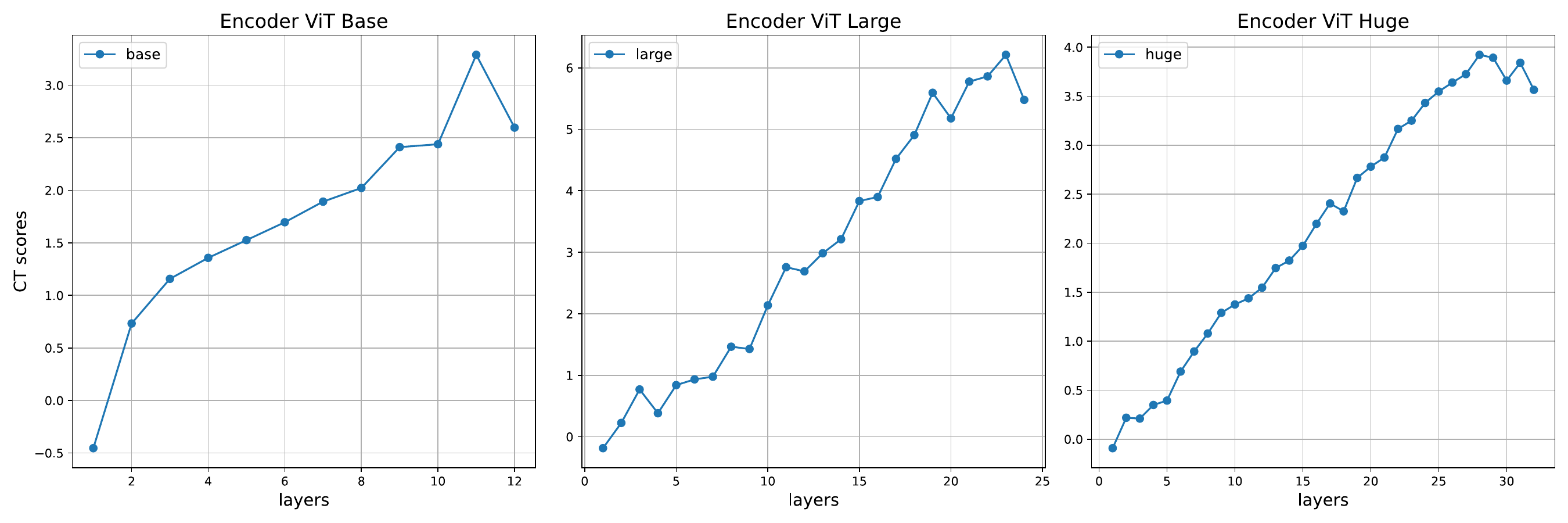}
\end{center}
\caption{\textbf{ViT encoder trend based on GAN-generated Image}. We use 500 random sampled BigGAN-deep \protect\citep{brock2019large} generated CIFAR-10 Images to compute the $C_T$ scores with three different Vision Transformers, namely ``vit-base-patch16'', ``vit-large-patch16'', and ``vit-huge-patch14.''}
\label{fig:GAN500ct}
\end{figure}

\begin{figure}[!ht]
\begin{center}

\includegraphics[width=\linewidth]{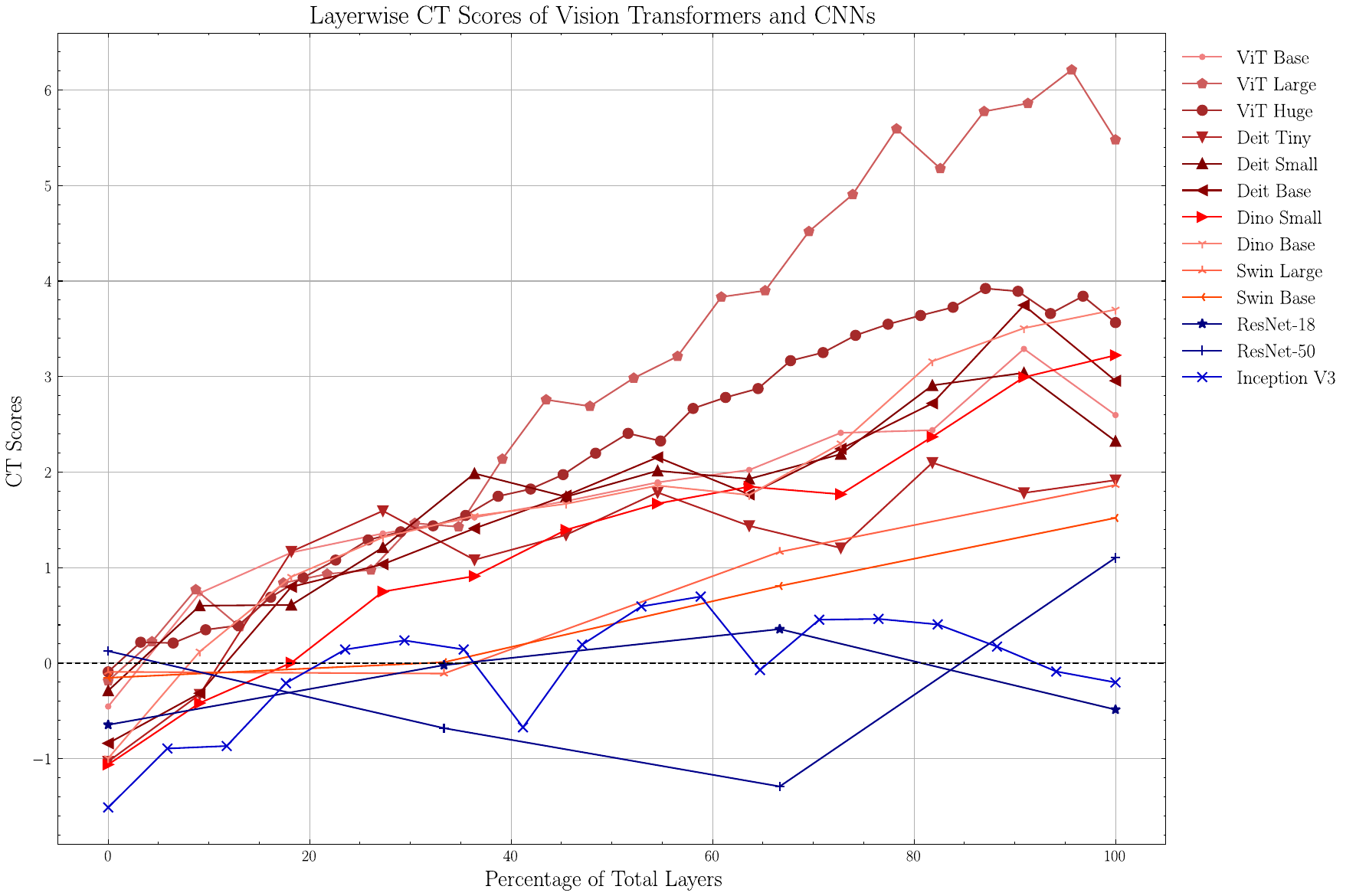}
\end{center}

\caption{\textbf{$C_T$ Score Comparison of ViT based Encoders and CNN Encoders}. This figure illustrates the trends in encoder performance for Vision Transformer, Dino ViT \protect\citep{caron2021emerging}, DeiT \protect\citep{touvron2021training}, and Swin Transformers \protect\citep{liu2021swin}, compared with ResNet \protect\citep{he2015deep} and InceptionV3 \protect\citep{heusel2018gans} encoders. Each graph is generated based on 500 GAN generated images, under similar settings to those in Figure \ref{fig:GAN500ct}.}
\label{fig:ViTCNNCompare}
\end{figure}

\subsection{What is $C_T$ Score}

The definition of $C_T$ score depends on the concept of data-copying. 
\citet{meehan2020nonparametric} defines that a generative model is data-copying the training set $T$ when in some regions, the model's output distribution $Q$ is systematically closer to $T$ than the true underlying distribution $P$, which $Q$ aims to learn. $C_T$-score is designed as a metric to assess the degree of data-copying by sampling and examining the generated data in relation to the training dataset ($T$) and test datasets (which is seen as samples from $P$). A higher $C_T$ score indicates a lower degree of data-copying and shows the model $Q$'s differentiation from $T$; whereas a lower score indicates a higher degree of data-copying and $Q$'s lack of originality.

Initially, embeddings for the training, test, and generated datasets are computed using pre-trained models, providing high-dimensional representations that capture the essence of the data's features. 
The embeddings are clustered using the KMeans algorithm, dividing the data embeddings into distinct cells that form a partition $\Pi$ of the instance space $\mathcal{X}$. This step is crucial for assessing data-copying behavior not only globally on the entire $\mathcal{X}$ but also locally within each cell.

Within each cell $\pi$, samples are drawn from the test and generated data embeddings. A distance function $d: \mathcal{X} \to \mathbb{R}$ is needed here to measure how close a sample is to the training set. For all the experiments in this paper, the nearest neighbor distance $d(x) = \min_{t \in T} ||x-t||^2$ is adopted. The samples $P_n$ (drawn from the test dataset) and $Q_m$ (drawn from the generated dataset) are passed into function $d$ and result in $L_{\pi}(Q_m)$ and $L_{\pi}(P_n)$, which are considered as samples from the distance distribution. Notice that the distance function $d$ serves as indicators of similarity, with shorter distances suggesting higher levels of similarity and potential copying.
Next, a $Z_U$ score is determined using the Mann-Whitney U test by comparing $L_{\pi}(Q_m)$ and $L_{\pi}(P_n)$ and quantifying their statistical difference. A higher $Z_u$ score indicates a greater disparity, implying less similarity and, consequently, less copying. Finally, the $C_T$ score is calculated as a weighted average of the $Z_U$ scores across all cells. The weighting is based on the proportion of the samples in each cell, and cells with an insufficient number of samples are excluded based on a predetermined threshold, ensuring the score only reflects cells with a meaningful representation of generated data. In short, the $C_t$ score can be formulated as

$$C_T(P_n, Q_m) = \frac{\sum_{\pi \in \Pi_{\tau}} P_n(\pi) Z_U(L_{\pi}(P_n), L_{\pi}(Q_m); T)}{\sum_{\pi \in \Pi_{\tau}} P_n(\pi)}$$

where $P_n(\pi)$ denotes the proportion and $\Pi_{\tau}$ denotes the set of cells with the number of samples above the threshold $\tau$. 

\subsection{Notable trend for ViT models}

Using the $C_T$-score formula provided, we conduct initial experiments to calculate layer-wise $C_T$-scores for encoders based on both transformer and CNN architectures.  Our findings reveal a unique trend observed in transformer encoders. $C_T$-scores are consistently increasing for latter layers, indicating lower memorization. 
This phenomenon is especially pronounced in the ViT architecture (as shown in Figure \ref{fig:ddpm500ct}, \ref{fig:GAN500ct}, \ref{fig:ViTCNNCompare} with different ViT implementations). It appears to be a characteristic feature of transformer-based encoders, suggesting a distinct pattern of information memorization associated with the layer hierarchy.

In contrast,  CNN-based encoders demonstrate a markedly different behavior. 
Our observations in Figure \ref{fig:ViTCNNCompare} indicate that CNNs exhibit a relatively consistent or flat $C_T$-score trend across their layers. 
This uniformity in $C_T$-score distribution suggests that CNN encoders process and represent data in a more homogeneous manner, distinguishing them as stable representers of information when compared to their transformer counterparts.
This stability suggests that, in the context of CNN encoders, the choice of layer for embedding extraction might not significantly impact memorization detection, echoing practices seen in other methodologies like Fréchet Inception Distance (FID) \citep{heusel2018gans} where models like InceptionV3 \citep{szegedy2015rethinking} are preferred for image embeddings extraction due to their consistency.

However, the intriguing results from the ViT models offer a contrasting perspective. 
Our observation of increasing $C_T$-scores across the layers of transformer encoders prompts a compelling hypothesis that different layers of the ViT model specialize in distinct levels of feature learning. In the context of images, the early layers of ViT concentrate on encoding low-level features, such as simple patterns, colors, and textures. In contrast, the latter layers focus on encoding high-level, more abstract features, such as object categories and complex scene relationships. This hypothesis necessitates a layer-specific analysis on using ViT encoders for embedding extractions, which is key to understanding how the choice of layers influences the memorization detection task.

\subsection{Low level and high level memorization}
\label{3.3 dummmy}

We hypothesize that generative models exhibit distinct patterns of memorization at both high and low levels of information, which are crucial for their functioning and output generation.
At the low level, these models fall into memorizing basic patterns, features, and textures. 
This form of memorization represents a limited understanding to elemental aspects such as edges, colors, and simple shapes. 
On the other hand, high-level memorization involves the integration and synthesis of these basic elements into more complex and abstract concepts. 
In visual models, this could mean recognizing and generating faces or landscapes. 
This high-level memorization is not merely a sum of low-level features but represents a more sophisticated semantic understanding and recombination.

In light of the intriguing trend of increasing $C_T$-scores observed in Vision Transformer (ViT) models, our study sought to establish a connection between this trend and the concepts of high-level and low-level memorization. We hypothesize that different layers within the ViT model capture distinct types of memorization representation.
Specifically, the embeddings from the initial (front/early) layers would primarily capture low-level memorization, focusing on fundamental features and patterns. The embeddings from the deeper (latter) layers would lean towards high-level memorization, characterized by the integration and synthesis of these basic elements into more complex and abstract representations.

\section{Experiment}

We design an experiment to measure the $C_T$-score trends using a Vision Transformer (ViT) encoder. Five datasets are created from the original CIFAR-10 using various augmentation techniques, which specifically perturb or preserve high-level or low-level information in the images. We expect to see a perturbation reflected in the early or latter parts of the $C_T$-Layer curve when the augmentation targets the low-level or high-level information, respectively. We use a BigGAN-deep model as the baseline model and the $C_T$-Layer curve calculated from its generation as the baseline curve to be compared against.

\subsection{Dataset preparation}

\begin{figure}[htbp]
\centering
\begin{subfigure}[b]{0.15\textwidth}
    \includegraphics[width=\textwidth]{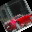}
    \caption{\scriptsize \textbf{Rotated}}
    \label{fig:datasetimage1}
\end{subfigure}
\begin{subfigure}[b]{0.15\textwidth}
    \includegraphics[width=\textwidth]{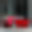}
    \caption{\scriptsize \textbf{Downsampled}}
    \label{fig:datasetimage2}
\end{subfigure}
\begin{subfigure}[b]{0.15\textwidth}
    \includegraphics[width=\textwidth]{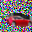}
    \caption{\scriptsize \textbf{Gaussian Seg}}
    \label{fig:datasetimage3}
\end{subfigure}
\begin{subfigure}[b]{0.15\textwidth}
    \includegraphics[width=\textwidth]{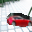}
    \caption{\scriptsize \textbf{Real Seg}}
    \label{fig:datasetimage4}
\end{subfigure}
\begin{subfigure}[b]{0.15\textwidth}
    \includegraphics[width=\textwidth]{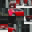}
    \caption{\scriptsize \textbf{Shuffled}}
    \label{fig:datasetimage5}
\end{subfigure}
\begin{subfigure}[b]{0.15\textwidth}
    \includegraphics[width=\textwidth]{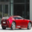}
    \caption{\scriptsize \textbf{CIFAR-10}}
    \label{fig:datasetimage6}
\end{subfigure}

\caption{Samples for curated datasets}
\label{fig:dataset_six_images}
\end{figure}

We curate five distinct datasets (examples: Figure \ref{fig:dataset_six_images}), which are augmented versions of CIFAR-10 to delve into the intricacies of low-level and high-level data processing mechanisms. Each modified version of the CIFAR-10 dataset (Figs. \ref{fig:datasetimage1}-\ref{fig:datasetimage5}) is treated as an output from ``pathological generative models,'' memorizing either low-level or high-level information disproportionately. Our objective is to ascertain whether modifications to either low-level or high-level features precipitate discernible alterations in the $C_T$-Layer curve, indicating the effect of such modifications. 
The deliberate modification of the CIFAR-10 dataset results in a high average dissimilarity between the augmented and original datasets. 
Consequently, the $C_T$-scores predominantly indicate underfitting (high positive values). Nevertheless, the relative trend observed in the $C_T$-Layer curve remains insightful as we are only interested in seeing shifts from the baseline consistently-increasing trend.

For the analysis of low-level features, we construct two datasets (Figs.\ref{fig:datasetimage1}-\ref{fig:datasetimage2}). We aim for these two datasets to imitate the generation of models that primarily alter low-level details while preserving high-level semantics. The first dataset (\ref{fig:datasetimage1}) is generated through the application of rotation-based modifications to CIFAR-10 images, incorporating a spectrum of angular orientations. 
The second dataset (\ref{fig:datasetimage2}) is constructed from the process of downsampling the original images followed by upsampling them back to their original dimensions.

Although the first two datasets intend to only perturb the low-level semantics, the processes of rotation and down-upsampling still corrupt the objects in the images to a degree, affecting the high-level information. To address this issue, the subsequent two datasets are crafted to imitate the generation of models that extract and preserve the training image's high-level semantics. In our exploration of such data processing techniques, we employ a fine-tuned ResNet-18 model (details of the fine-tuning process are provided in Appendix \ref{sec:Appendix:1}) to conduct image segmentation on images from the CIFAR-10 dataset.
This segmentation isolates the foreground elements of the images, which are then seamlessly integrated onto alternative backgrounds.
Specifically, the modification entails replacing the original image backgrounds with Gaussian noise \ref{fig:datasetimage3} and real-world landscapes \ref{fig:datasetimage4}, sourced from the BG-20k \citep{li2021bridging} dataset. 
This strategic modification is designed to ensure that the fundamental semantics of the image—primarily its main object—remain largely unaffected.

Following the four datasets that preserve high-level information while altering low-level details, we construct an additional dataset (\ref{fig:datasetimage5}) that largely preserves the low-level details while significantly alters the high-level information. This dataset is generated by fragmenting each original CIFAR-10 image into 16 equal-sized square patches and shuffling them. Each image in this dataset has the exact collection of pixels, though arranged differently, as its original counterpart. This can be viewed as an intentional preservation of the low-level image features since color and pattern details within each piece are exactly the same as in the original.

\begin{figure}[h]
\centering

\begin{subfigure}[b]{0.32\textwidth}
    \includegraphics[width=\textwidth]{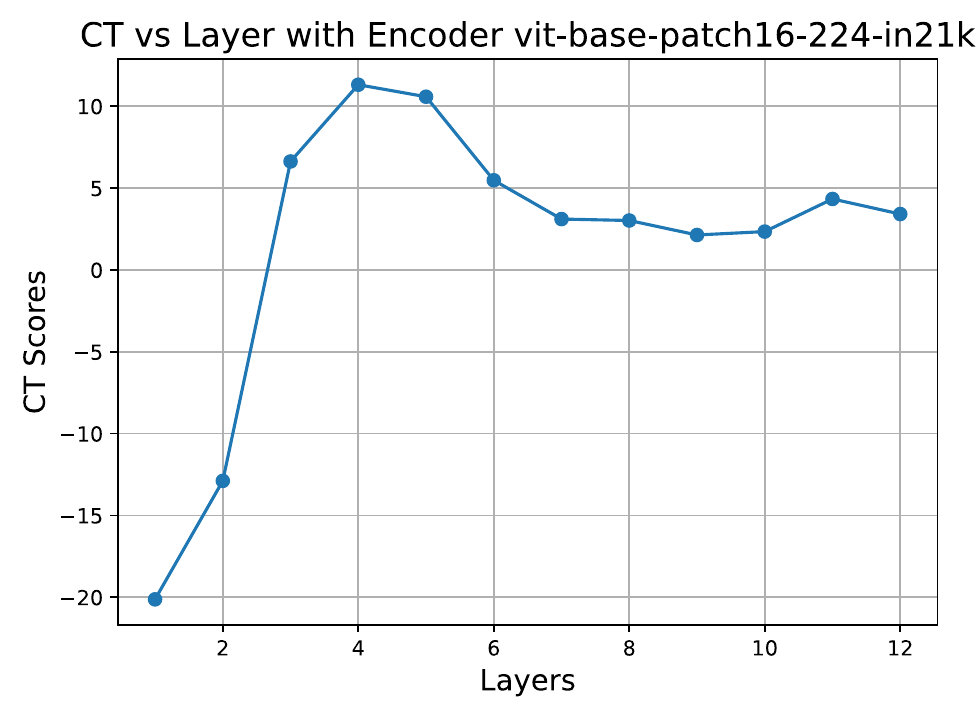}
    \caption{\textbf{Rotated}}
    \label{fig:ctdatasetimage1}
    \includegraphics[width=\textwidth]{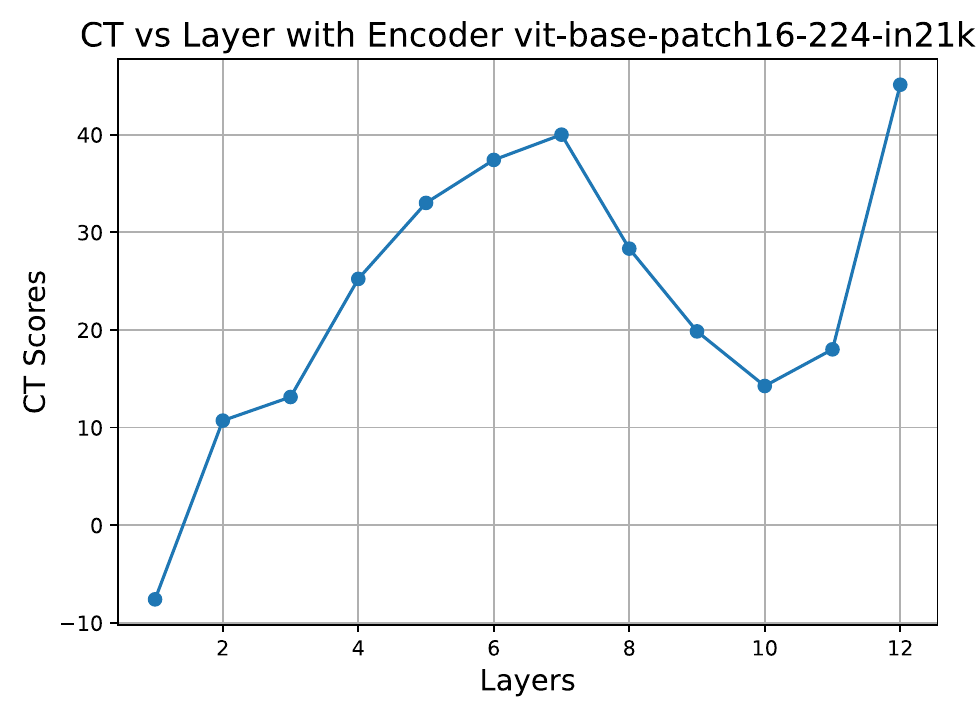}
    \caption{\textbf{Downsampled}}
    \label{fig:ctdatasetimage2}
\end{subfigure}
\hfill 
\begin{subfigure}[b]{0.32\textwidth}
    \includegraphics[width=\textwidth]{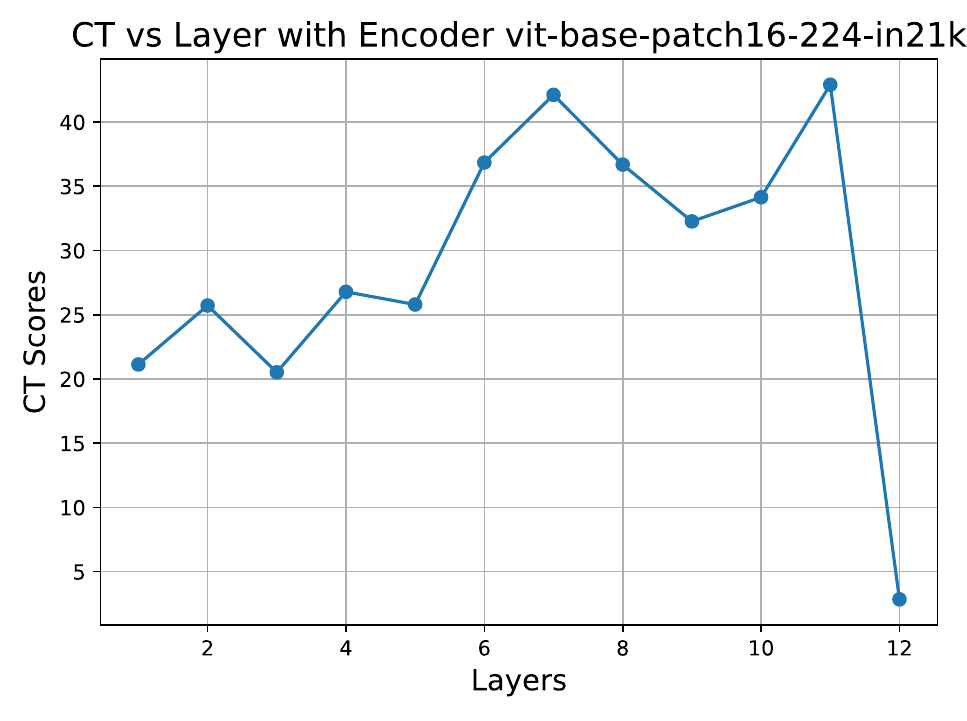}
    \caption{\textbf{Gaussian Seg}}
    \label{fig:ctdatasetimage3}
    \includegraphics[width=\textwidth]{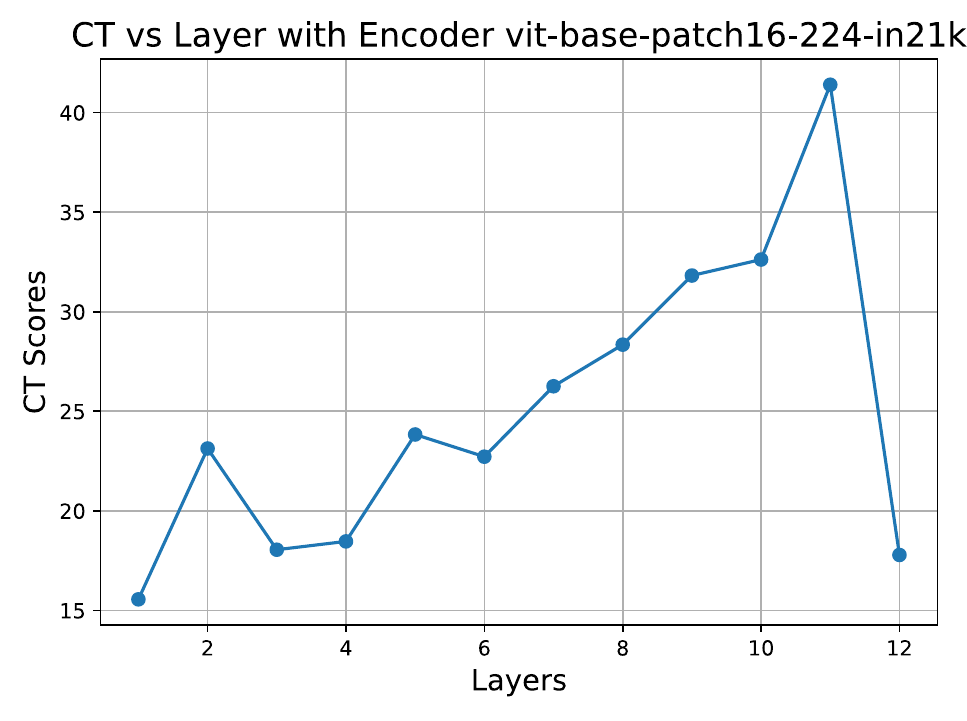}
    \caption{\textbf{Real Seg}}
    \label{fig:ctdatasetimage4}
\end{subfigure}
\hfill
\begin{subfigure}[b]{0.32\textwidth}
    \includegraphics[width=\textwidth]{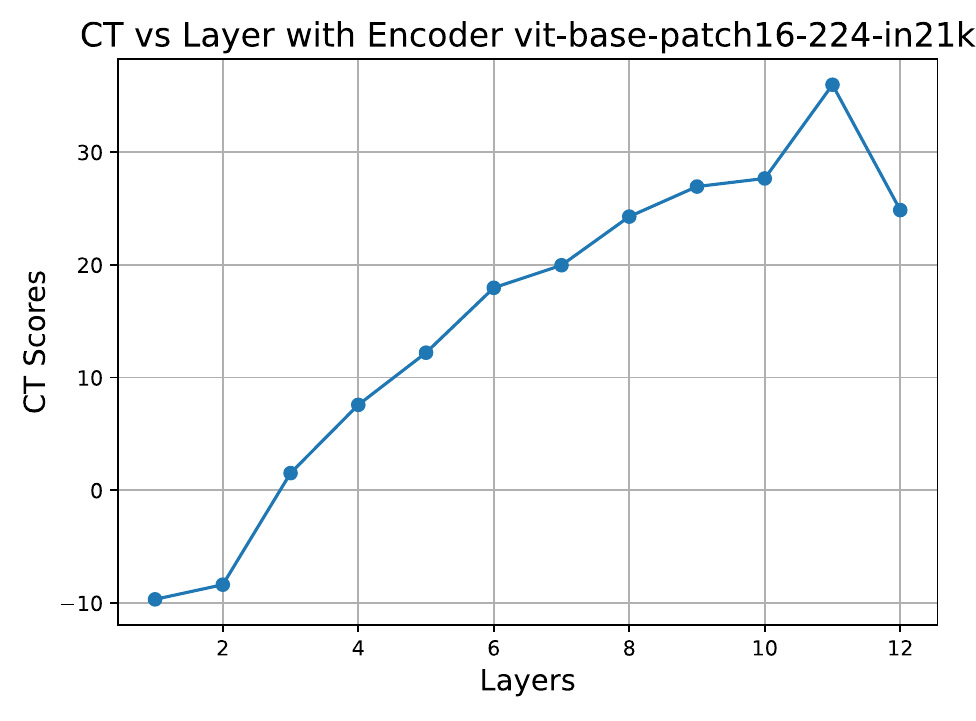}
    \caption{\textbf{Shuffled}}
    \label{fig:ctdatasetimage5}
    \includegraphics[width=\textwidth]{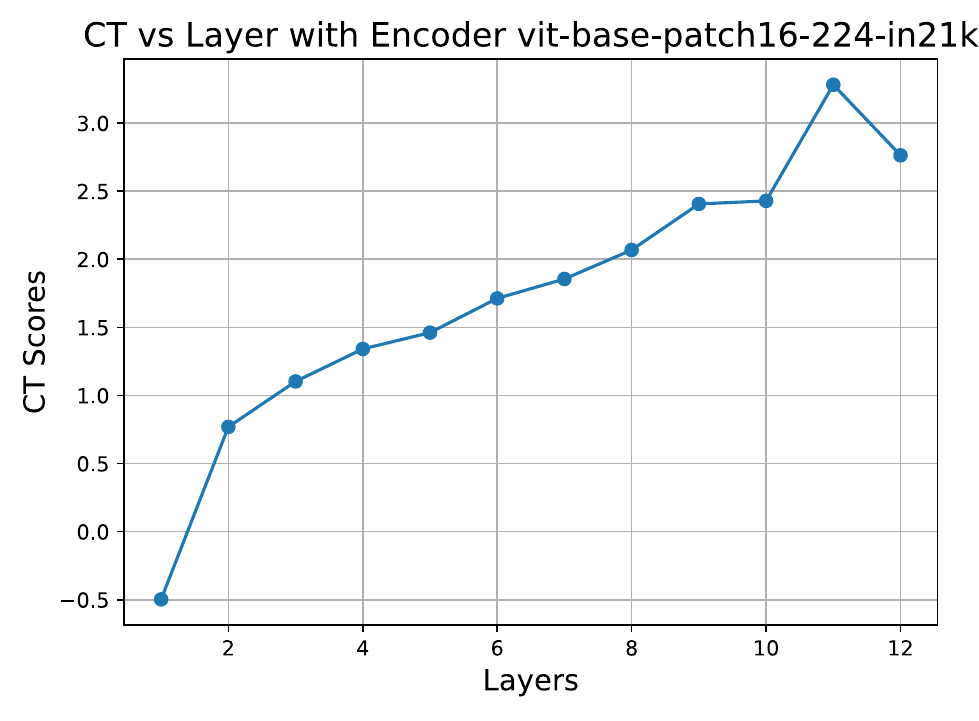}
    \caption{\textbf{Reference (GAN)}}
    \label{fig:ctdatasetimage6}
\end{subfigure}

\caption{$C_T$ vs Layers Curve for Six different dataset}
\label{fig:ctdataset_six_images}
\end{figure}

\subsection{$C_T$-Scores Analysis}

We employ a non-pathological generative model (specifically,  BigGAN-deep model in Fig.\ref{fig:GAN500ct} ) as a baseline model, and use a ViT-base model as the encoder to derive the $C_T$-layer curves for all six models (Fig.\ref{fig:ctdataset_six_images}). The observed shifts of the augmented dataset $C_T$-layer trends from the baseline curve closely align with our hypothesis in \ref{3.3 dummmy} that embeddings from initial layers capture low-level memorization, while those the from deeper layers exhibit high-level memorization.

For datasets augmented with low-level detail modifications -- rotation and resolution changes (\ref{fig:ctdatasetimage1}, \ref{fig:ctdatasetimage2}) -- we observe a significant shift in their $C_T$-layer trends at the initial layers, compared with the baseline increasing trend (\ref{fig:ctdatasetimage6}). The $C_T$-layer curves for both rotation and down-upsampling display an early peak at the initial layers. Since these two datasets are more dissimilar to the original CIFAR-10 in terms of low-level details,
this observation underscores the sensitivity of the network's earlier layers to alterations in low-level image details.

On the other hand, the analysis of high-level detail preservation, achieved through the alteration of image backgrounds while preserving the foreground elements, reveals the latter network layers' focus on high-level semantic information. There are discernible drops of the relative $C_T$-scores on the last layer of the network for the two segmentation methods (\ref{fig:ctdatasetimage3}, \ref{fig:ctdatasetimage4}), which both intend to cause overfitting on high-level information by preserving the semantics of original CIFAR-10.

In the analysis of the $C_T$-scores curve for the shuffled dataset (\ref{fig:ctdatasetimage5}), the observed deviation is relatively subtle, attributable to the inherently increasing trend of the baseline reference. Consequently, even an enhancement in the scores of the deeper layers aligns with this prevailing trend. Notably, the stability in the score of the front layers, despite significant alterations to high-level details, substantiates the hypothesis that the deeper layers are chiefly tasked with the extraction of high-level semantic information from the input images.

\subsection{Further Experiment}
Our first experiment analyzes layer-wise $C_T$-scores trends on reflecting the memorization of low-level and high-level information. It supports our hypothesis in \ref{3.3 dummmy} on the difference in memorization detection between different layers of encoder embeddings. It heavily relies on the assumption that all the specially augmented images are samples from some imaginary ``pathological generative models.'' In contrast, the next experiment uses samples from real trained generative models of two specific architectures - denoising diffusion probabilistic models (DDPM) \citep{ho2020denoising} and denoising diffusion implicit models (DDIM) \citep{song2022denoisingdiffusionimplicitmodels}. DDPM and and DDIM are both Diffusion Models that generate data distributions from noises by adding noise to data over time steps and then learning to reverse the process. DDPM follows strict Markovian assumption at inference time, while DDIM follows a deterministic non-Markovian formulation. We're interested in seeing if layer-wise $C_T$-scores can capture the memorization characteristics of the two models. 

\begin{figure}[h]
\centering
\textbf{}
\begin{subfigure}[b]{0.58\textwidth}
    \includegraphics[width=\textwidth]{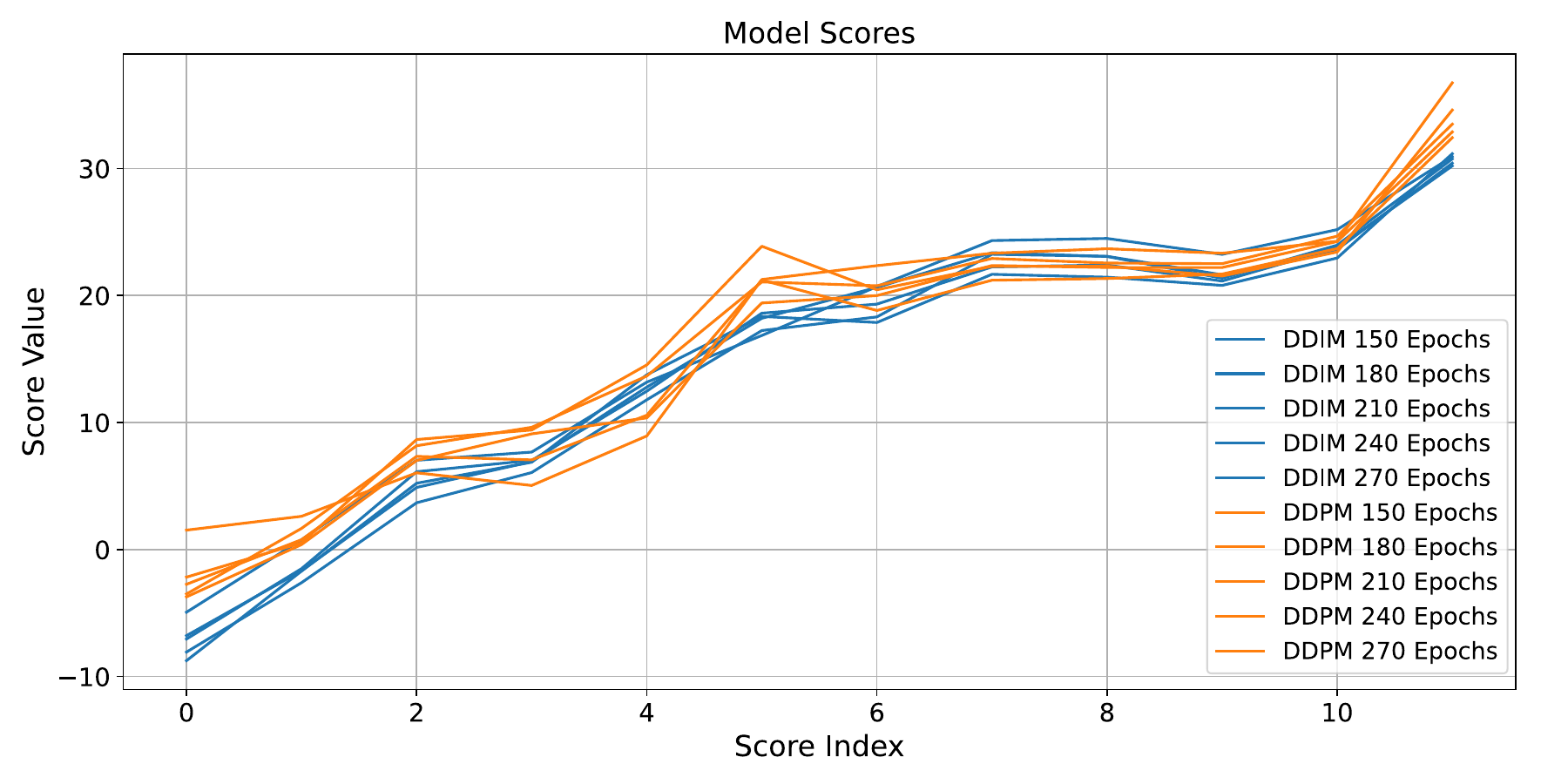}
    \caption{\textbf{Raw $C_T$-Layer curves}}
    \label{fig:DDPM&DDIM_RAW}
\end{subfigure}
\hfill 
\begin{subfigure}[b]{0.4\textwidth}
    \includegraphics[width=\textwidth]{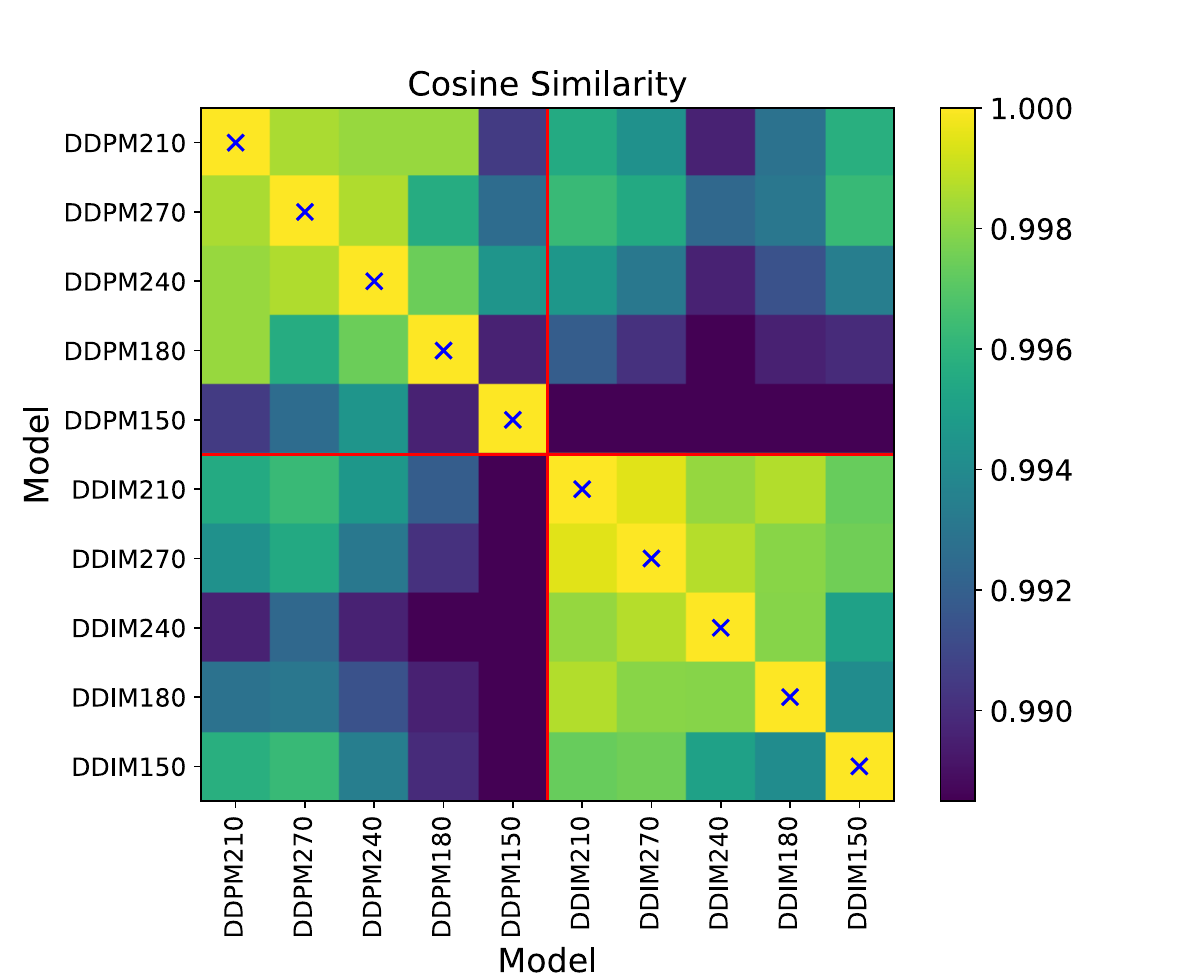}
    \caption{\textbf{Pair-wise Cosine Similarities}}
    \label{fig:heatmap}
\end{subfigure}
\caption{\textbf{$C_T$-scores of Multiple Checkpoints of DDPM and DDIM} (a) The raw $C_T$-Layer curves of checkpoints when training epochs = 150, 180, 210, 240, 270 for both DDPM and DDIM (b) 
The heat map records all pair-wise cosine similarities between checkpoints. The heat map is symmetric, with the 1st and 3rd quadrants displaying inter-architecture similarities, and the 2nd and 4th quadrants displaying intra-architecture similarities. The diagonal is crossed out since diagonal blocks correspond to all the checkpoints' similarities with themselves (always 1).
} 
\label{fig:TRAINING_SIZE}
\end{figure}

For robustness purpose, multiple checkpoints are taken at several epochs for both DDPM and DDIM (details of the training are provided in Appendix \ref{sec:Appendix:2}). Only checkpoints at epochs after the loss function largely converges are recorded to ensure the image generation is meaningful. In this case, epoch number 120, 150, 180, 210, 240, and 270 are used. The layer-wise $C_T$-Layer scores curves \ref{fig:DDPM&DDIM_RAW} show a similar increasing trends for both architectures on all checkpoints. This meets our expectation because it's unlikely that the added number of training epochs will change the model memorization natures drastically. The distinction between DDPM and DDIM architectures might not be prominent due to the fact that they adopt very similar diffusion methods. 

However, the subtle divergence between the overall DDPM trend (orange curves) and DDIM trend (blue curves) is still noticeable. This makes it intriguing to compare the intra-architecture (e.g. DDPM180 and DDPM210) and inter-architecture (e.g. DDPM180 and DDIM180) $C_T$-Layer curve similarities. By treating each $C_T$-Layer curve as a vector in $\mathbb{R}^{11}$, where 11 is the number of encoder layers, the cosine similarity between each pair of $C_T$-Layer curves is computed and recorded in a heat map \ref{fig:heatmap}. The overall intra-architecture similarity is higher than the inter-architecture similarity, even if the raw curves in \ref{fig:DDPM&DDIM_RAW} look fairly similar. This experiment result suggests that the memorization characteristics are unique to model architectures, not to specific training checkpoints.

\section{Fingerprinting}
Our findings indicate that the $C_T$-Layer curve serves as a meaningful indicator of the model's handling of different levels of details, suggesting that this curve could act as a distinctive feature reflective of the model. This insight paves the way for utilizing the $C_T$-Layer curve as a technique for fingerprinting generative models, an application growing in importance within the field as people are more and more worried about deepfake and malicious use of generated media.

\subsection{Introduction to Model Fingerprinting Using $C_T$-Layer Curves}

\begin{figure}[ht]
\centering
\includegraphics[width=0.9\textwidth]{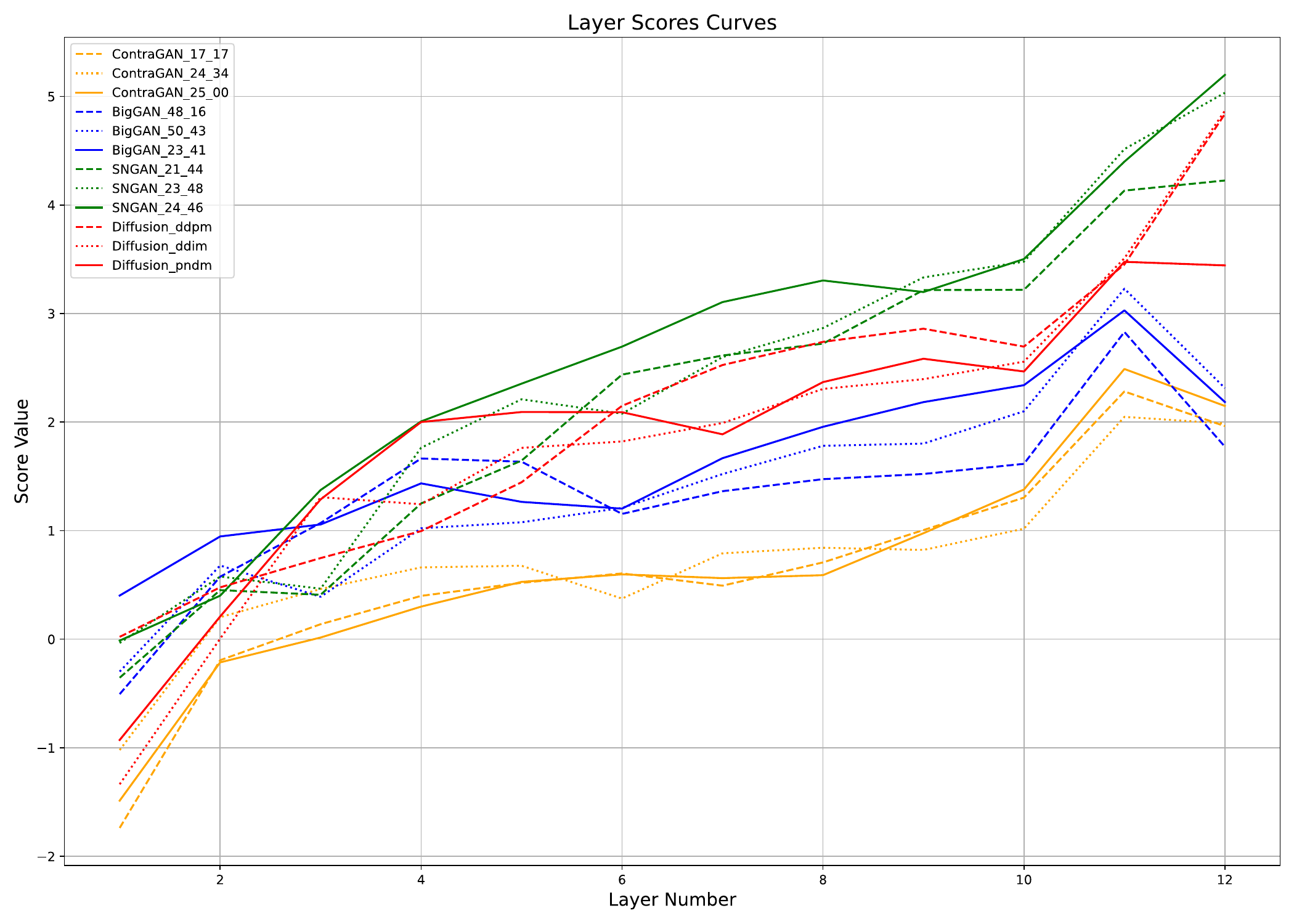}
\caption{\textbf{Differentiation of Model Architectures Utilizing $C_T$-Layer Scores Curves.} The displayed graph highlights distinct trajectories for each model, revealing the utility of $C_T$ Layers curves in differentiating between various model architectures. The convergence of models with the same architecture into identifiable clusters further reinforces the curves' efficacy as a discriminative tool. This demonstrates the curves' potential as a reliable method for architectural distinction and analysis in model evaluation.}
\label{fig:ct_curves}
\end{figure}

Traditional methods of fingerprinting generated images often rely on access to the processed training images \citep{yu2022artificial} utilized by the model during its training phase, a requirement that is not always practical or feasible. In contrast, our proposed methodology offers a novel approach to extract characteristic features of a given model architecture by analyzing its corresponding $C_T$-Layer scores \ref{fig:ct_curves}. For the purpose of this study, we employ the ViT-base model as our encoder; However, the methodology is adaptable and could potentially be extended to other variations of ViT encoders.

In more detail, our method involves calculating the $C_T$ score for each layer of the target image set. Subsequently, we employ these scores in conjunction with L2-norm and cosine-similarity nearest neighbor analysis to identify the closest match within our dataset. This dataset comprises $C_T$-Layer scores for a diverse array of known models, enabling the identification and matching process. Through this approach, we introduce a robust mechanism for model fingerprinting that bypasses the need for direct access to a model's training images, thereby offering a more versatile and accessible method for characterizing and identifying generative models.

\subsection{Experimental Setup for Evaluating Fingerprinting Accuracy}

To empirically validate our proposed fingerprinting methodology, we conduct a preliminary experiment encompassing a diverse set of 18 model samples derived from various generative model architectures. The architectures are ContraGAN, BigGAN, and SNGAN for GANs and DDPM, DDIM, and PNDM for Diffusion Models. For each architecture, we have three model samples, each generating 500 images, resulting in a total of 1,500 generated images per architecture.

To establish a comparative framework, we implement a baseline method that utilizes the mean, standard deviation (std), and Fréchet Inception Distance (FID) score to extract features from the generated images and perform nearest-neighbor matching. Additionally, we compare our approach with several neural network classification methods, including a vanilla CNN model trained from scratch, a fine-tuned ViT-base model, and a fine-tuned ResNet50 model (details of the training and fine-tuning process are provided in Appendix \ref{sec:Appendix:3}).

Our experimental setup is meticulously crafted to evaluate the accuracy of our method. From the 1,500 generated images for each model architecture, we first separate 500 images as the testing set. The remaining 1,000 images are then used to form two training datasets (Datasets\_small with 500 images and Datasets\_full with the entire 1,000 images) to create ablation studies. For our $C_T$ method, we formulate the training-set $C_T$-scores for each model using its corresponding training images. At testing time, for each target, we calculate its $C_T$-score using the testing set and identify the best match in the training-set $C_T$-scores with different matching methods. The overall accuracy is determined by the proportion of correctly matched target models. For the baseline method, the overall process mirrors that of the $C_T$ method, but we use baseline statistical scores instead. 

We train different neural networks on training sets and evaluate their accuracy on the testing set. We use majority votes for final classification since each model takes a single image as the input, while our $C_T$ and baseline methods are applied to the entire testing dataset. By employing this systematic approach, we aim to provide a thorough and robust assessment of our proposed method's performance relative to both statistics-based and neural network-based techniques.

\begin{table}[h]
\centering
\small 
\resizebox{0.9\textwidth}{!}{%
\begin{tabular}{|l|c|c|c|c|}
\hline
\multicolumn{1}{|c|}{\textbf{Method}} & \multicolumn{2}{c|}{\textbf{ Dataset (small)}} & \multicolumn{2}{c|}{\textbf{ Dataset (full)}} \\
\cline{2-5}
 & \textbf{Accuracy} & \textbf{Time(s)} & \textbf{Accuracy} & \textbf{Time(s)} \\
\hline
Baseline Method (L2 Norm) & 0.666 & \textbf{22.46} & 0.833 & \textbf{44.73} \\
Baseline Method (Cosine Similarity) & 0.333 & \textbf{22.46} & 0.333 & \textbf{44.73} \\
Baseline Method (L2 Norm + Cosine Similarity) & 0.666 & \textbf{22.46} & 0.833 & \textbf{44.73} \\
Vanilla CNN Model & 0.500 & 200 & 0.833 & 400 \\
ViT-base Model & 0.167 & 1830 & 0.333 & 3642 \\
ResNet50 Model & \textbf{0.833} & 1130 & \textbf{1.000} & 2278 \\
\textbf{Our:} $C_T$-Layer Method (L2 Norm) & 0.666 & 1417 & \textbf{1.000} & 2835 \\
\textbf{Our:} $C_T$-Layer Method (Cosine Similarity) & 0.666 & 1417 & 0.833 & 2835 \\
\textbf{Our:} $C_T$-Layer Method (L2 Norm + Cosine Similarity) & \textbf{0.833} & 1417 & \textbf{1.000} & 2835 \\
\hline
\end{tabular}%
}
\caption{Comparison of Baseline Model and $C_T$-Layer Method with Time Metrics}
\label{tab:fingerprint_comparison}
\end{table}

\subsection{Results and Performance Analysis}

The results outlined in Table \ref{tab:fingerprint_comparison} underscore the effectiveness of our proposed $C_T$-layer methodology. Compared to the baseline methods, our approach offers a significant improvement in performance, demonstrating its capability to capture and leverage architectural nuances for model fingerprinting. The $C_T$-Layer Method records a remarkable increase in accuracy for both the full and small sets, with an outstanding 100\% accuracy in the full set and an 83.3\% accuracy in the small set using the combination of L2 Norm and Cosine Similarity. This performance significantly outstrips the baseline methods, which peak at 66.6\% and 83.3\% accuracy for similar conditions. 

Additionally, our methodology outperforms conventional models such as Vanilla CNN and ViT-base, and it achieves comparable results to ResNet50, while without the need for training. Although our method employs a ViT-base model as the encoder, direct finetuning a ViT-base model to learn fingerprinting classification yields suboptimal results due to limited data availability — a common challenge in fingerprinting scenarios.

Moreover, we evaluate the training duration of our method against other models. The baseline method is exceptionally quick due to its simplistic statistical nature. However, our approach demonstrates comparable processing times to ResNet50 and is faster than the ViT-base model, underscoring its efficiency.

These results suggest that the $C_T$-layer scores, by leveraging layer-specific responses to various image details, provide an effective method for model identification and characterization. This improvement is critical in scenarios where precision and reliability in model fingerprinting are paramount. That the $C_T$-Layer Method outperforms the established models confirms its efficacy and potential for generative model fingerprinting.

\section{Conclusion}

This study pioneers an approach to understanding data memorization in generative models, providing a detailed analysis of the impact of layer selection in Convolutional Neural Networks (CNNs) and Transformers on memorization detection. It reveals distinct memorization patterns between CNNs and Vision Transformers (ViTs), underscoring the importance of layer-specific analysis for accurate memorization quantification. Furthermore, the introduction of a novel model fingerprinting technique leveraging $C_T$-score distributions marks a significant advancement in identifying generative models, offering potential tools for the ethical use of generative models, and addressing critical concerns around privacy, content integrity, and the proliferation of deepfakes.

\paragraph{Limitations and future directions}

While our proposed technique offers significant advancements, several practical limitations and challenges need to be considered for real-world applications.

The choice of layer for memorization detection varies depending on the type of memorization (low-level vs. high-level), which can be challenging to distinguish without extensive analysis. Future work should develop hybrid detection methods that analyze both low-level and high-level features and automate layer selection algorithms for better accuracy. Although our fingerprinting method does not require direct access to the model's training dataset, it still necessitates a general, compact, and sufficiently generic dataset to serve as a practical baseline. Future directions should include curating such datasets and implementing dynamic updates to include new and emerging generative model characteristics.

Our method requires pre-computed $C_T$-Layer score data for various generative model architectures. Future efforts should focus on expanding this database to include new models and establishing collaborations with developers for early access. The computational complexity of computing $C_T$-Layer scores for each layer can be intensive. Future research should aim to optimize computation processes and leverage high-performance computing resources to improve efficiency and scalability.

Implementing our fingerprinting technique requires careful consideration of ethical and privacy implications. Establishing clear guidelines and ethical standards for deployment, ensuring data privacy, and developing protocols for transparency and accountability will be crucial. Collaborations with ethicists, policymakers, and industry stakeholders will help align technological advancements with societal values and norms.

\newpage
\bibliography{main}
\bibliographystyle{tmlr}

\newpage
\appendix
\section*{Appendix}
\renewcommand\thesubsection{\Alph{subsection}}
\label{sec:Appendix}

\subsection{Fine-Tuning the ResNet-18 Model for Image Segmentation}
\label{sec:Appendix:1}

This section provides details on the fine-tuning process of the ResNet-18 model for image segmentation tasks on the CIFAR-10 dataset.

We start with a pre-trained ResNet-18 model from PyTorch \citet{paszke2019pytorchimperativestylehighperformance} and modify its architecture to suit the image segmentation task based on \citet{CharlieLehman_2019}'s implementation. The first convolutional layer is replaced to adapt to CIFAR-10 image dimensions. The final fully connected layers are removed, and a bilinear upsampling layer is added to increase the resolution of feature maps. Additionally, a convolutional layer for classification with 10 output channels (corresponding to the CIFAR-10 classes) is appended, followed by another bilinear upsampling layer to match the output size to the input size. The CIFAR-10 dataset is used for both training and testing. For training, the images are transformed using random cropping, horizontal flipping, normalization, and conversion to tensors. For testing, the images are normalized and converted to tensors without any augmentation.

The training procedure involves setting up the loss function, optimizer, and learning rate scheduler. The model is trained for 30 epochs with the following details:

\begin{itemize}
    \item \textbf{Loss Function:} Binary Cross-Entropy with Logits Loss (BCEWithLogitsLoss)
    \item \textbf{Optimizer:} Stochastic Gradient Descent (SGD) with a learning rate of 0.05, momentum of 0.9, and weight decay of 1e-4
    \item \textbf{Learning Rate Scheduler:} Cosine Annealing with a minimum learning rate of 0.001
    \item \textbf{Batch Size:} 128 for training, 100 for testing
    \item \textbf{Number of Epochs:} 30
    \item \textbf{Number of Workers:} 16 for data loading
\end{itemize}

The training and evaluation of the model are conducted on a machine equipped with an NVIDIA GeForce GTX 1080 Ti GPU. This computational environment facilitates the efficient processing of the CIFAR-10 dataset and the fine-tuning of the ResNet-18 model.

During training, both training and validation loss and accuracy are monitored to evaluate the model's performance. These metrics are plotted to visualize the model's learning progress over the epochs. The segmentation results are also visualized by converting the segmented outputs to images, highlighting the model's ability to segment different classes within the CIFAR-10 dataset.

\begin{table}[ht!]
\centering
\begin{tabular}{|l|l|l|}
\hline
\textbf{Method} & DDPM & DDIM\\ \hline
\textbf{Noise-predicting Model Architecture} & \multicolumn{2}{c|} {Unet2D} \\ \hline
\textbf{Loss Function} & MSE Loss & L1 Loss \\ \hline
\textbf{Training Time Step} & \multicolumn{2}{c|} {1000} \\ \hline
\textbf{Inference Time Step} & 1000 & 250 \\ \hline
\textbf{Training Batch Size} & \multicolumn{2}{c|} {16} \\ \hline
\textbf{Optimizer} & \multicolumn{2}{c|} {AdamW (LR: 0.0001)} \\ \hline
\textbf{Device} & \multicolumn{2}{c|}{NVIDIA GeForce GTX 1080 Ti GPU} \\ \hline
\textbf{Training Dataset} & \multicolumn{2}{c|} {CIFAR-10 (Resolution: 32)} \\ \hline
\end{tabular}
\caption{Summary of DDPM and DDIM Method Configurations and Training Settings}
\label{tab:DDPM-DDIM}
\end{table}

\subsection{Training DDIM and DDPM Models}
\label{sec:Appendix:2}

This section details training DDIM and DDPM models and collecting their checkpoints at multiple epochs.

The DDPM model implementation follows the original DDPM paper \citep{ho2020denoising}. We use the official training and inference code posted by the authors on the Hugging Face website. The DDIM implementation follows a GitHub Repository: \citet{simple-diffusion-git}. Table~\ref{tab:DDPM-DDIM} summarizes the details of training and inference configurations of the two methods.

\subsection{Baseline Neural Network Models for Fingerprinting Performance Comparison}
\label{sec:Appendix:3}

This section provides details on the baseline neural network models used for performance comparison in the context of fingerprinting tasks. Specifically, we describe the training of a vanilla Convolutional Neural Network (CNN) model, a fine-tuned ResNet-50 model, and a fine-tuned Vision Transformer (ViT) model from PyTorch \citet{paszke2019pytorchimperativestylehighperformance}.

The table below summarizes the architecture, loss function, optimizer, and other training settings for the three models:

\begin{table}[th!]
\centering
\begin{tabular}{|l|l|l|l|}
\hline
\textbf{Model} & Vanilla CNN & ResNet-50 & Vision Transformer (ViT) \\ \hline
\textbf{Architecture} & 3 Conv layers, 2 FC layers & Pre-trained, modified FC layer & Pre-trained, modified head \\ \hline
\textbf{Loss Function} & \multicolumn{3}{c|} {Cross-Entropy Loss} \\ \hline
\textbf{Optimizer} & Adam (LR: 0.001) & Adam (LR: 0.001) & Adam (LR: 0.01) \\ \hline
\textbf{Epochs} & 50 & 50 & 30 \\ \hline
\textbf{Batch Size} & 32 & 32 & 32 \\ \hline
\textbf{Device} & \multicolumn{3}{c|}{NVIDIA GeForce GTX 1080 Ti GPU} \\ \hline
\textbf{Dataset} & \multicolumn{3}{c|}{Fingerprinting dataset with resize to 224x224 (ResNet-50 and ViT)} \\ \hline
\textbf{Transforms} & \multicolumn{3}{c|}{Resize, Normalize} \\ \hline
\end{tabular}
\caption{Summary of Model Architectures and Training Settings for Fingerprinting}
\end{table}

The performance of the models is evaluated using accuracy metrics. During training, the average loss and accuracy per epoch are monitored. After training, a majority vote mechanism is used for evaluation on the test dataset. Predictions for each label are collected, and the most common prediction is considered as the final prediction for that label. The accuracy is calculated based on the correct majority votes, providing insight into the models' classification capabilities for the fingerprinting task.

\end{document}